%% file: SDE-DiffusionMap.tex
\documentclass{article}
\usepackage{Style-SDE-DiffusionMap}
\pdfoutput=1
\usepackage{subfiles}

\title{Diffusion Maps for Signal Filtering in Graph Learning}
\author{Todd Hildebrant}
\date{December 2023}

\begin{document}

\maketitle

\begin{abstract}
This paper explores the application diffusion maps as graph shift operators in understanding the underlying geometry of graph signals. The study evaluates the improvements in graph learning when using diffusion map generated filters to the Markov Variation minimization problem. The paper showcases the effectiveness of this approach through examples involving synthetically generated and real-world temperature sensor data. These examples also compare the diffusion map graph signal model with other commonly used graph signal operators. The results provide new approaches for the analysis and understanding of complex, non-Euclidean data structures.
\end{abstract}

\section{Introduction}
\subfile{sections/Intro-SDE-DiffusionMap}

\section{Background}
\subfile{sections/Background-SDE-DiffusionMap}

\section{Applying Diffusion Maps to the Graph Shift Operator}

\subfile{sections/Applying-SDE-DiffusionMap}

\section{Examples}
\subfile{sections/Examples-SDE-DiffusionMap}

\section{Conclusion}

\subfile{sections/Conclusion-SDE-DiffusionMap}

\newpage
\section{Appendix} \label{Appendix}

\subfile{sections/Appendix-SDE-DiffusionMap}

\newpage
\printbibliography

\end{document}

%% file: sections/Intro-SDE-DiffusionMap.tex
Diffusion Maps, introduced by Coifman and Lafon in 2006\cite{coifmanDiffusionMaps2006}\footnote{ 
"We use the eigenfunctions of a Markov matrix that define a random walk on the data to obtain new descriptions of the data sets (subset of $\mathbb{R}^n$, graphs) via a family of mappings that we term 'diffusion maps.' These mappings embed the data points into a Euclidean space in which the usual distance describes the relationship between pairs of points in terms of their connectivity. This defines a useful distance between points in the data set that we term the diffusion distance.'" 
\cite{coifmanDiffusionMaps2006} 
}, 
have proven to be a versatile tool for nonlinear dimension reduction and for capturing the underlying geometry of complex data sets. Similarly, graph signal processing has become a framework for analyzing signals residing on irregular graph structures. Combining these two methodologies allows a more nuanced understanding of data exhibiting geometric and graph-based characteristics. In contemporary data analysis, analysts face increasingly complex data requirements. Some challenges exhibit non-Euclidean characteristics, such as finding superior representation in graphs that excel at capturing intricate relationships. GSP integrates algebraic and spectral graph-theoretic concepts with computational harmonic analysis to facilitate signal processing directly on the graph. These techniques have found diverse applications in big data analysis \cite{mateosConnectingDotsIdentifying2019}, power grid state forecasting and power grid cyberattack detection \cite{wuComplexValueSpatiotemporalGraph2022}, and brain network analysis \cite{dongGraphSignalProcessing2020}.

\subsection{Dimension Reduction}

When studying complex data, a challenge is how to decrease the complexity of the data set while preserving relevant information for clustering, classification, and regression. Dimension reduction aims to strike a balance between complexity and clarity. Linear dimension reduction techniques, such as Principal Component Analysis (PCA), identify orthogonal axes (principal components) that capture data variability. Linear dimension reduction is computationally efficient and provides understandable results.

Similarly, spectral clustering uses the eigenvectors and eigenvalues of a pairwise similarity matrix for geometric data analysis. Spectral clustering techniques, discussed in \cite{ngSpectralClusteringAnalysis} and \cite{CHUNG1997}, represent a graph using a matrix for calculating the eigenvalues and eigenvectors. The Laplacian plays a crucial role, and different distance measures impact the modeling. Spectral clustering views the data as a spectral space through the eigenvalues and eigenvectors of the Laplacian matrix.

Recently, attention has shifted to kernel eigen-map techniques. These nonlinear techniques aim to capture complex relationships in high-dimensional data. Nonlinear-dimensional reduction techniques, such as t-distributed Stochastic Neighbor Embedding (t-SNE), Locally Linear Embedding (LLE), and Isomap, emphasize the preservation of local structures. However, they can be computationally intensive and sensitive to parameter choices. Graph coarsening, or graph summarization, reduces a large graph to a smaller one while preserving properties. \cite{loukasStationaryTimevertexSignal2019} developed frameworks for graph matrix coarsening that maintain spectral and cut guarantees, focusing on the graph adjacency matrix. Diffusion maps use the Markov matrix to illustrate the connections between data points. They create a transition probability matrix using a Gaussian kernel function, capturing the main modes of diffusion and providing a reduced-dimensional representation by eigenvalue and eigenvector decomposition.

Diffusion Maps are used to reduce the dimensionality of data by embedding them in a lower-dimensional space so that the Euclidean distance between points approximates the diffusion distance in the original feature space. This process creates the diffusion distances, and then uses a diffusion map to reduce the dimensionality and approximate the manifold's distances non-linearly. The features of the diffusion map allow for a lower-dimensional analysis. Diffusion Maps operate by modeling the diffusion process on a given dataset. By constructing a Markov transition matrix that encodes pairwise similarities between data points, the diffusion map captures the global geometry of the data in a low-dimensional space. The eigenfunctions of this matrix provide a set of coordinates that preserve the intrinsic structure of the data.

\subsection{Graph Signal Processing}

GSP extends signal processing to signals on non-Euclidean domains represented by weighted graphs. Signals on graphs are attribute values associated with each node. GSP broadens classical signal processing methods for irregular graph-structured data. Graph Signal Processing extends classical signal processing to signals residing on irregular graph structures. It leverages graph Laplacians and Fourier transforms to analyze and process signals defined on graphs. GSP provides a framework for understanding the spectral properties of graph signals, enabling the development of advanced filtering and transformation techniques.

The field of Graph Signal Processing (GSP) was made famous by \cite{shumanEmergingFieldSignal2013}, \cite{sandryhailaDiscreteSignalProcessing2013}, \cite{talmonDiffusionMapsSignal2013}, and \cite{ortegaGraphSignalProcessing2018}. These works approached the problem from two distinct perspectives. The authors of \cite{shumanEmergingFieldSignal2013}, based their definition on the graph Laplacian and elements from harmonic analysis, leading to the rationalization of frequencies and filter banks in the graph domain. On the other hand, the authors of \cite{sandryhailaDiscreteSignalProcessing2013} followed a construction similar to algebraic signal processing and defined their fundamental concept, the graph shift operator. One valid extension is the design of filters from polynomials of the graph shift operator, enabling operations such as smoothing, denoising, and feature extraction. Spectral analysis and techniques like Laplacian eigenmaps are used in GSP for dimensional reduction and embedding graph-structured data into lower-dimensional spaces.

In classical data analysis and signal processing, the signal domain is typically determined by equidistant time instants or a set of spatial sensing points on a uniform grid. A more recent approach, graph signal processing, allows for a more general domain, where a signal is characterized by assigning real (or complex) data values, \(x(n)\), to each vertex of a graph. Graph signal processing originated from two different approaches: one based on the spectral representation of the graph and the other based on algebraic signal processing. In \cite{sandryhailaDiscreteSignalProcessing2013}, following the algebraic signal processing construction, the authors first identified the DSP time shift operator, $z^{-1}$, as the most straightforward non-trivial operator which can delay the input signal $x$ by one sample. They then multiply a graph signal $x$ by the graph's adjacency matrix  $A$ and show this serves as the graph time shift operator $Ax = \tilde{x}$. \cite{leusGraphSignalProcessing2023} then offers how the two approaches can be combined by defining a generic 'graph shift operator' (GSO) that can be viewed as the most basic operation applied to a graph signal, and it then describes the graph more generally way than $L$ or $A$. This flexibility allows the linear GSO $S \in \mathbb{R}^{N \times N}$ to have a variety of constructions based on the application. However, care must be taken as the choice of shift operator leads to different signal models, so shift matrixes that are diagonalizable or not defective are usually considered.

\subsection{Summary of this Paper}

This paper begins with the view that a diffusion map is a Graph Shift Operator(GSO) and that filtering the signals using this GSO both reduces the dimensionality of the data and removes noise. This GSO is then used to process graph signals to improve the results when learning the graph structure. Then, examples are given to show the approach's effectiveness on randomly generated data and actual temperature sensor data \cite{molenedata} through the graph learning problem and compares the results with other commonly used GSOs by measuring the similarities of the graphs.

\subsection{Related Work} 

\cite{lafonDiffusionMapsCoarsegraining2006} was the first paper to use the diffusion map for graph coarsening, or dimension reduction of a graph that ensures that a sub-graph has the same spectral characteristics as the original. \cite{heimowitzUnifiedViewDiffusion2017} was the first paper to use the Markov matrix as a graph shift operator and then mention using the diffusion map to reduce dimensions. They define a graph filter based on this shift operator but do not pursue the idea further. \cite{eliasDiffusionbasedVirtualGraph2020} uses the Markov relationships between graph nodes to extend the adjacency matrix by calculating the probability of connecting the nodes. \cite{gasteigerDiffusionImprovesGraph2022} uses transition probabilities and the diffusion process to create a graph diffusion convolution to further optimize certain graph neural network (GNN) problems. Additional analysis of graph shift operators is given in \cite{gaviliShiftOperatorGraph2017} and \cite{roddenberrySamplingLimitTheories2023}. To the best of my knowledge, no other work uses the diffusion map as a graph shift operator to generate a signal model. Furthermore, using this operator in graph learning is new.

\subsection{Common Notation}

Graphs in this paper are restricted to simple undirected weighted graphs with nonnegative edge weights and no self-loops. A graph $\mathcal{G}( \mathcal{V}, \mathcal{E} )$ has $(m$ nodes and edges $\mathcal{V}$. The neighborhood $\mathcal{N}(\mathcal{V'})$ of a given collection of nodes $\mathcal{V'} \subseteq \mathcal{V}$ is the set of nodes $\mathcal{V'}$ and those nodes immediately connected to it, and is defined as $ \mathcal{N}(\mathcal{V}) = \{ u \in \mathcal{V} : (u,v) \in \mathcal{E}$ for some  $v \in \mathcal{V'} \} \cup \mathcal{V'}$. For a given graph $\mathcal{G}$, we can define the space of graph signals as $\mathcal{X}(\mathcal{G}) = \{x: \mathcal{V} \rightarrow \mathbb{R}$. Usually, the sampled data is represented as a matrix $X \in \mathbb{R} ^ {m \times n} = [x_1,x_2,...,x_n]^\mathsf{T}$ where $m$ are the variables of interest, and $n$ is the number of observations of features. The weights $\mathcal{W}$ of the graph store the distances between vertices in $\mathcal{V}$ connected by the edges in $\mathcal{E}$. The connectivity of the graph is represented by the adjacency matrix $A \in \{0,1\}^{ m \times m}$ or the weight matrix $W \in \mathbb{R}^{m \times m}$.

%% file: sections/Background-SDE-DiffusionMap.tex
\subsection{Diffusion Maps} 

Diffusion Maps are based on an analogy to the heat diffusion process and begin by constructing a transition matrix to represent the probability that a particle will move to an adjacent position in the next time step. More specifically, given a measure space $(X, A, \mu)$, where $X$ is the data set, and $\mu$ is the distribution of points in $X$, we define $d(x) = \int_X k(x, y) \, d\mu(y)$ as a local measure. $k(x,y)$ is a symmetric kernel with $k(x, y) = k(y, x)$ and $k(x, y) \geq 0$. Thus, the probability of moving from $x$ to $y$  is $p(x, y) = \frac{k(x, y)}{d(x)}$ which is the transition kernel of a Markov chain on $X$. For $t > 0$, the probability of transitioning from $x$ to $y$ in $t$ time steps is given by $P^t(x, y)$, the transition kernel of a Markov chain on $X$, which is the kernel of the $t$-th power $P^t$ of $P$.

A data-centric formulation is to take a set of data $\mathbf{X}$ arranged in a matrix, with each column $\mathbf{x}_i$ representing a data point in a high-dimensional space. Create a graph $\mathcal{G}$ by first determining the pairwise distance $D$ between data points by $D_{ij} = \| \mathbf{x}_i - \mathbf{x}_j\| ^2$. Choose a kernel $k$ that is appropriate for the application\footnote{while the Gaussian kernel is widely used, both for ease of computation as well as its association with the heat equation, other kernel choices include the Friis transmission equation and the log-distance path loss model in wireless applications. \cite{ghafourianWirelessLocalizationDiffusion2020}} and create the weighted adjacency matrix $A_{i,j}$. A common choice of the Gaussian kernel can be expressed as  
\[  
A_{i,j} =\begin{cases} \exp\left(-\frac{D_{ij}}{2\sigma^2}\right)& \text{if \(i,j\) is \( \in \mathcal{E}\) },\\ 0& \text{if not} \end{cases} 
\]  
where $\sigma$ is a bandwidth parameter. Details about how to choose $\sigma$ are in \ref{Appendix}. The resulting adjacency matrix is symmetric and semi-definite.

Next, create the degree matrix $D$ with diagonal elements $D_{ii} = \sum_{j} A_{ij}$, and then form the transition matrix (also known as the random walk transition matrix, or Markov matrix) $P$  by $P = D^{-1} A$.  The eigenvectors of $P$ are $\psi ^N_{i=1}$ with corresponding eigenvalues that form a discrete sequence $\{\lambda_l\}_{l \geq 0}$ and such that $1 = \lambda_0 > |\lambda_1| \geq |\lambda_2| \geq \ldots \geq |\lambda_N|$. Since the spectrum of the data matrix has a spectral gap with only a few eigenvalues close to one and all additional eigenvalues much smaller than one, we can use only the first few eigenvectors. For example, figure \ref{fig:eigendropoff} shows the eigenvalue decay for the later Molene Temperature example. The diffusion maps are then 

\[ \Psi _t (i) = \begin{bmatrix}  
\lambda^t _2 \psi _2 (i) \\ 
\lambda^t _3 \psi _3 (i) \\ 
\dots \\ 
\lambda^t _N \psi _N (i) 
\end{bmatrix}, \quad 
\text{for } i = 1 \dots N. 
\] 

Each $\Psi_t(x)$ element is a diffusion coordinate. The map $\Psi_t: X \to \mathbb{R}^{N,t}$ embeds the data set into a Euclidean space such that the distance in this space is equivalent to the diffusion distance  
\[
D ^2 _t (x, y) = \ \Psi_t(x) - \Psi_t(y)\ = \sum_{m=2} ^N \lambda _m ^{2t} ( \psi _m (x) - \psi _m (y))^2. 
\] 
The mapping  $\Psi_t: X \to \mathbb{R}^{N,t}$ provides a parameterization of the data set or, equivalently, a realization of the graph G as a cloud of points in a lower-dimensional space $\mathbb{R}$ where the rescaled eigenvectors are the coordinates. The dimensionality reduction and the weighting of the relevant eigenvectors are dictated by both the time t of the random walk and the spectral fall-off of the eigenvalues. The above equation means that the embedding is done in such a way that the Euclidean distance approximates the diffusion distance. Since the eigenvalues decrease rapidly, most of the information in the diffusion map is represented in the first few $\lambda$ values. This allows for a reduction in dimension to $l << N$ and the final embedding is a matrix with the first $l$ eigenvalue-scaled eigenvectors as columns. The rows of this matrix are the coordinates of the graph vertices in the reduced dimension.

However, $P$ is not necessarily symmetric, which might result in complex eigenvectors. So, similar to \cite{delaporteIntroductionDiffusionMaps2008}, first create the symmetric transition matrix $P_{sym} = D^{\frac{1}{2}} P D^{- \frac{1}{2}}$ which has the same eivenvalues as $P$, and eigenvectors which transformed as $\psi_P = D^{-\frac{1}{2}} \psi_{P_{sym}}$. The diffusion maps are created using \ref{DM-process}. 

\begin{algorithm}\label{DM-process} 

\caption{Diffusion Map generation} 

\begin{algorithmic} 

\Require Data Set $X \in \mathbb{R}^{n \times k}$ $n$ observations of $k$ elements and $\sigma$ 

\State Define a kernel $d(x,y)$ and construct a weight matrix $W(i,j) = exp (- \frac{d(X_i,X_j)^2}{2 \sigma ^2})$ 

\State Compute $D$ the degree matrix with a diagonal equal to the row sums of $W$ 

\State Compute the diffusion matrix $P = D^{-1} W$ 

\State Compute the symmetric diffusion matrix $P_{sym} = D^{\frac{1}{2}} P D^{- \frac{1}{2}}$ 

\State Compute the eigenvalues and eigenvectors of $P_{sym}$ 

\State $\lambda_{P_{sym}} = \lambda_P$ and $\psi_P = D^{-\frac{1}{2}} \psi_{P_{sym}}$ 

\State Construct the diffusion map $\Psi$ as $\lambda \psi$ arranging the top $l << N$ as columns 

\end{algorithmic} 

\end{algorithm}

\subsection{Graph Signal Processing}

In DSP, generic linear filters are constructed as polynomials $p(z)$ of consecutive applications of $z^{-1}$ and are shift invariant $z_{-1}p(z) = p(z) z^{-1}$. \cite{sandryhailaDiscreteSignalProcessing2013} also defines linear shift-invariant filters and shows that they are polynomials in the graph shift operator $H(x) = h(A)x = (h_0 I + h_1 A + \dots + h_L A L)x$. Now, with the GSO $S$, and if $V$ is the matrix containing in its columns the eigenvectors of $S$, then linear graph filtering can be understood equivalently as $\hat{x} = V^{-1} x$. A graph filter $H(x)$, a polynomial of order $L$, is mixing values that are at most $L$ hops away, the polynomial coefficients $h_l$ representing the strength given to each of the neighborhoods. This representation now allows for more complicated filters to be built based on the polynomial or the GSO. For example, (this is also a low pass graph filter) Tikhonov regularization $h(x) = \frac{1}{1+\alpha x}$, or smoothing using heat diffusion, $h(x) = exp(-t x)$ \cite{kalofoliasHowLearnGraph2016}.

Several approaches exist to define an equivalence of the Fourier transform. In all of them, the graph Fourier transform is a linear operator and can be represented by a matrix. The Graph Fourier Transform of a signal $\mathbf{x}$ is the column vector representing the signal $x$, and $\mathbf{U}$ is the matrix with columns as the eigenvectors $u_i$, then the GFT is given by $\hat{\mathbf{x}} = \mathbf{U}^\top \mathbf{x}$. The graph Fourier transform diagonalizes the graph shift operator and provides a way to analyze the graph signals in the spectral domain. The inverse Graph Fourier Transform can be used to reconstruct the signal $x$ from its Fourier coefficients $x(v) = \sum_{i=0}^{n-1} \hat{x}(\lambda_i) u_i(v)$, where $u_i(v)$ is the value of the $i$-th eigenvector at the vertex $v$. In matrix form $\mathbf{x} = \mathbf{U} \hat{\mathbf{x}}$, where $\mathbf{U}$ is the matrix with its columns the eigenvectors $u_i$. From the GFT of the signal, common SP concepts can now be defined in GSP, including ordering graph frequencies or design of low- and high-pass graph filters.

GSP was initially developed with a known graph of relationships. Still, in many applications, the graph is an implicit object that describes the connections or levels of association between the variables. In some cases, the graph links can be based on expert domain knowledge (e.g., activation properties in protein-to-protein networks). However, in many other cases, the graph must be inferred from the data. This problem can be formulated as: given a collection of M graph signals $X = [x_1, \ldots, x_M] \in \mathbb{R}^{N \times M}$, find an $N \times N$ sparse graph matrix describing the relations among the graph nodes. \cite{leusGraphSignalProcessing2023}

The early approaches focused on the information associated with each node separately so that the existence of the link $(i, j)$ in the graph was decided based only on the $i$-th and $j$-th rows of $X$. GSP contributes to the problem of graph learning by providing a more sophisticated (spectral or polynomial) way to relate the signals and the graph.

Most graph learning problems are in two categories. One attempt to discover information about the underlying geometry of the graph is by inferring relationships between the signal and the associated nodes. The other uses information about the graph's geometry to understand aspects of the signals and finds optimal sampling or interpolation strategies. A group of early GSP works focused on learning a graph that made the signals smooth with respect to the learned graph. If smoothness is promoted using Laplacian-based total variation, the formulation leads to a regression problem with the pseudoinverse of $L$ as the kernel.

A second set of GSP-based topology inference methods models the data resulting from a diffusion process on the sought graph through a graph filter. Assumptions (if any) about the diffusing filter $H(x)$ and the input signals $x_m$ are vital questions when modeling observations. Assuming the signals \(x_m\) are stationary leads to a model in which the covariance (precision) matrix of the observations is a polynomial of the GSO \(S\).

%% file: sections/Applying-SDE-DiffusionMap.tex
\subsection{Graph Shift Operator}

By focusing on analogies with the classical signal processing time shift or delay operator as a fundamental operation that shifts a signal along the time axis without altering its shape, the graph shift operator (GSO) was initially defined in \cite{sandryhailaDiscreteSignalProcessing2013} as a matrix $S \in \mathbb{R}^{N \times N}$ with $S_{ij} = 0$ for $i \neq j$ and $i, i \notin \mathcal{E}$. This means that $S$ can take nonzero values at the edges of $\mathcal{G}$ or its diagonal, although this is not required. $A$ \cite{shumanEmergingFieldSignal2013} and $L$ \cite{sandryhailaDiscreteSignalProcessing2013} were used in the development of GSP, and \cite{ortegaGraphSignalProcessing2018} later showed them to be equivilent. \cite{heimowitzUnifiedViewDiffusion2017} showed how $P$ can be used as a GSO, allowing one to define a graph signal as a function of diffusion embedding vectors instead of a function over the nodes of a graph. Another approach in \cite{shumanEmergingFieldSignal2013} observes that the translation of a signal x is the convolution of x with the Dirac mass $\delta_\tau$. In recent years, the GSO has been reframed as an algebraic descriptor of the graph topology in \cite{mateosConnectingDotsIdentifying2019} and extended to consider energy preservation shifts \cite{gaviliShiftOperatorGraph2017} and isometric shifts \cite{giraultTranslationGraphsIsometric2015}.

Expanding on the original definition, the current definition of a GSO $S$, associated with a graph structure $\mathcal{G}$, is a linear mapping between graph signals $S: \mathcal{X}(\mathcal{G}) \rightarrow \mathcal{X}(\mathcal{G})$ such that the output graph signal $y = Sx$ is calculated as $y_v = \sum_{v \in \mathcal{N}(v)} S_{uv} X_u$. The GSO is determined by the adjacency rule that assigns the values $S_{uv}$ for every $(u,v) \in \mathcal{E}$. \cite{giraultSignalProcessingGraphs} summarizes three properties\footnote{The paper also defines Localization preserving: $\forall i \exists k_i: S (\delta_i) = \delta_{k_i}$ (delta signal is mapped to another delta signal) as another property. But, the paper then shows that all four properties of the time shift can be preserved if and only if the graph Fourier transform is a DFT, which is unlikely, so there is no usable operator on graph signals verifying all four properties. Thus, they choose to enforce only the first three: linearity, convolution, and isometric concerning the chosen energy definition.} that a graph shift operator should share with the time shift:  

\begin{itemize} 

    \item Linear: $S(\alpha x + y) = \alpha S(x) + S(y)$ (superposition principle) 

    \item Convolutive: $\forall v_k \neq v_l, \quad (\hat{S})_{kl} = 0$ 

    \item Energy preserving (isometric): $E_x = E_{S(x)}; E_x = x \times x = \sum_{i \in \mathcal{V}} |x_i|^2$  

\end{itemize} 

Additionally, to ensure numeric stability, we assume that the spectral norm of the shift operator is upper bounded, i.e., $\| S\| _2 \leq \rho$.

In \cite{heimowitzUnifiedViewDiffusion2017}, the authors proposed using the Markov matrix as a graph shift operator, which would redistribute the graph signal according to the transition probabilities, so, for node $v_i$, the shift operation would result in $\sum_{m=1}^N P(v_i v_m) x_m$. Realizing that, according to the diffusion map process, most of the graph's geometry is described by the largest eigenvalues and their associated eigenvectors of the Markov matrix, we can take the next logical step and consider the diffusion map embedding as a graph shift operator.

The diffusion map process can also act as a filter for graph signals, with the benefit of reducing complexity \cite{talmonDiffusionMapsSignal2013}. Take the data set and compute the diffusion maps $\Psi_t(i)$ as before. Since the eigenvectors are orthogonal, they form the complete basis for any actual signal, allowing it to be viewed as $x = \sum_{j=0}^{M-1} \lambda_j \langle x(j)\Psi_j \rangle \Psi_j$. Therefore, a subset of the eigenvectors $l << N$ exists that conveys the signal information, with the remainder representing noise. This can be expressed as a graph filter $h(x)_l$, a linear projection onto the eigenvectors spanning the parameter subspace.

The primary benefit of this definition is that it allows us to define a graph signal as a function of the diffusion embedding vectors instead of a function of the nodes of a graph since the diffusion map embedding preserves the local structure of the graph. Thus, we consider the diffusion map embedding of the original data as a graph shift operator, $S_{DM}$, defined as $S_{DM} = \Psi_P = \lambda_P \psi_P$. Intuitively, this embedding has the same information as the transition matrix $P$ but in a lower dimension and can detect temporal patterns in the graph. The diffusion map framework establishes an embedding for the graph nodes contingent on the timescale parameter $t$. Consequently, $S_{DM}$ is closely related to a time shift and can be employed to identify temporal patterns in the graph, expressed as  

\[ 
x = V\hat{x} = \Psi_0\hat{x}, \quad P^\top x = V\Lambda^\top \hat{x}= \Psi_T\hat{x}. 
\]
This means that the graph shift operation can be interpreted as a shift in the family of embedding, for example, from $\{ \Psi_0 \}$ to $\{ \Psi_T \}$. \cite{heimowitzUnifiedViewDiffusion2017} One additional assumption is that the graph signals are stationary with respect to a specific GSO $S$ if either the signal can be expressed as the output of a graph filter with white inputs or if its covariance matrix is simultaneously diagonalizable with S. \cite{giraultStationaryGraphSignals2015}

\begin{theorem} 

    $S_{DM}$ is a GSO 

\end{theorem}

\begin{figure}[h] 

    \centering 

    \includegraphics[scale=.5]{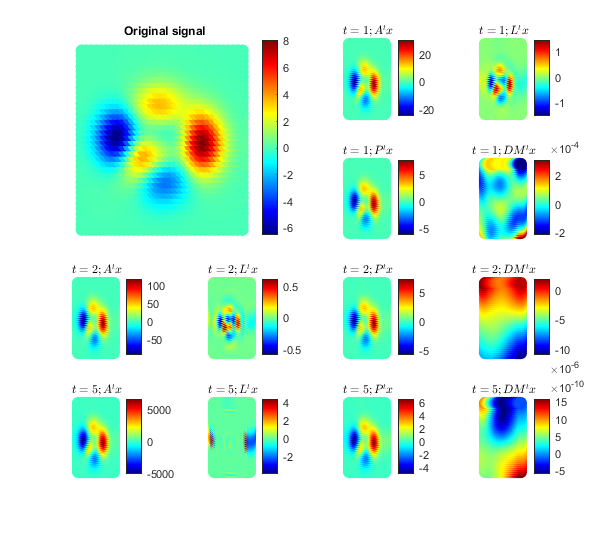} 

    \caption{applying different $H(S)^t$ to $x$} 

    \label{fig:gso_examples} 

\end{figure}

\begin{proof} 

To determine if $S_{DM}$ is a GSO, apply the elements of the definition: 

\begin{itemize} 

    \item The symmetry of $S_{DM}$ depends on P. If P was generated using a symmetric distance function, P is symmetric. Alternatively, use \ref{DM-process} to construct $S_{DM}$ based on $L_{sym}$.  

    \item Symmetry implies real eigenvalues and orthogonal real eigenvectors, so $S_{DM}$ is real. 

    \item $\| S_{DM} \| = \sqrt{\lambda_1}$ where $\lambda_1$ is the largest eigenvalue of $S_{DM}$. $S_{DM}=\Psi_P = \lambda_P \psi_P$. Since P is a Markov matrix, it is row stochastic, and $\rho(P) = 1$, so $| \lambda | \leq 1$. Thus $\lambda_1 \leq 1$ so $\|S_SM\|$ is bounded.  

    \item For $x,y \in \mathcal{X}$, $S_{DM}(\alpha x + y) = \Psi_i (\alpha x_i + y_i) = \alpha \Psi_i x_i + \Psi_i y_i$ since $\Psi_i = \lambda_i \psi_i =$ is constant. 

    \item  The convolutive property refers to the ability of $\hat{S_{DM}}$ to have an orthogonal diagonalization. Since $S_{DM}$ is real, symmetric, and isometric, then $F = \Psi(S_{DM})$ is orthogonal, so $\hat{S_{DM}} = F S_{DM} F^{-1}$ and then $\forall l: v_k \neq v_i, \quad \hat{S_{DM}}_{ki} = 0$  

    \item Isometric matrices have orthogonal columns, resulting from $S_{DM}$ being real. Furthermore, since P is stochastic, we have $D(P)=1$. As $S_{DM} = \Psi \Lambda \Psi^{-1}$, applying the graph shift operator $n$ times to the graph signal $x$ results in $x_n = S_{DM}^n x = \Psi \Lambda^n \Psi^{-1} x \approx \pi x$ because $\pi = P \pi = \frac{\psi_0}{\sum_i \psi_i}$. Therefore, $S_{DM}$ preserves the energy of the signal $x$. 
\end{itemize} 
\end{proof} 
The graph shift operator $S_{DM}$ can be approximated using diffusion maps as $S_{DM}^t \approx I - P^t$, as the diffusion operator $P^t$ is related to the graph Laplacian by formula $P^t = \exp(-tL)$, where $t$ is diffusion time. Thus, $S_{DM}$ represents one discrete step in how information moves from one node to its neighbors. In figure \ref{fig:gso_examples}, different simple filters $H(S)$ are applied to a simple signal $\delta(i) = 1$ associated with an equally weighted 2D-lattice graph. Repeated iterations show how the signal spreads based on the adjoining nodes. Since the kernel function ensures that closely connected nodes in the graph have close diffusion embedding vectors, the graph signal on the vertices is mapped onto the set of diffusion embedding vectors. This suggests that band-limited graph signals can approximate the graph signal when the graph shift operator is the Markov matrix. By enforcing the preservation of the power spectrum (i.e., through isometry and convolution), we allowed the graph translation to preserve the bandwidth of the signal through multiple applications of the operator.  \cite{gaviliShiftOperatorGraph2017} 

Since the choice of the GSO determines the signal model, using $S_{DM}$, we can now write the general graph filter as  

\[ 
H (x) = h_0\Psi_0\hat{x} + h_1\Psi_1\hat{x} + \dots + h_L\Psi_L \hat{x}. 
\] 

Usually, $l << N$, so $H(x)$ will be a low-degree matrix polynomial, and the output will strictly depend on the connectivity structure and signal values in the node's local neighborhood. As $S_{DM} = \Psi \Lambda \Psi ^{\top}$ The GFT is now given by $\hat{\mathbf{x}} = \mathbf{\Psi}^\top \mathbf{x}$. The iGFT is $\mathbf{x} = \mathbf{\Psi} \hat{\mathbf{x}}$. In general, filtering a graph signal, $x \in \mathbb{R}^m$ by $H$ is defined as 

\[  
y = H(S_{DM})x = \sum_i u_i h(\lambda_i) \hat{x} 
\]  

where $ \{u_i, \lambda_i \}$ are the pairs of eigenvalues-eigenvectors of $S_{DM}$, and $\hat{x} \in \mathbb{R}^m$ is the graph Fourier representation of $x$. The convolution operator is now $(f * g)(i) = \sum_{l=0}^{N-1} \hat{f}(\lambda_l)\hat{g}(\lambda_l)u_l(i)$, so convolution in the vertex domain is equal to multiplication in the graph spectral domain. 

\begin{figure}[h] 

    \centering 

    \includegraphics[scale=.5]{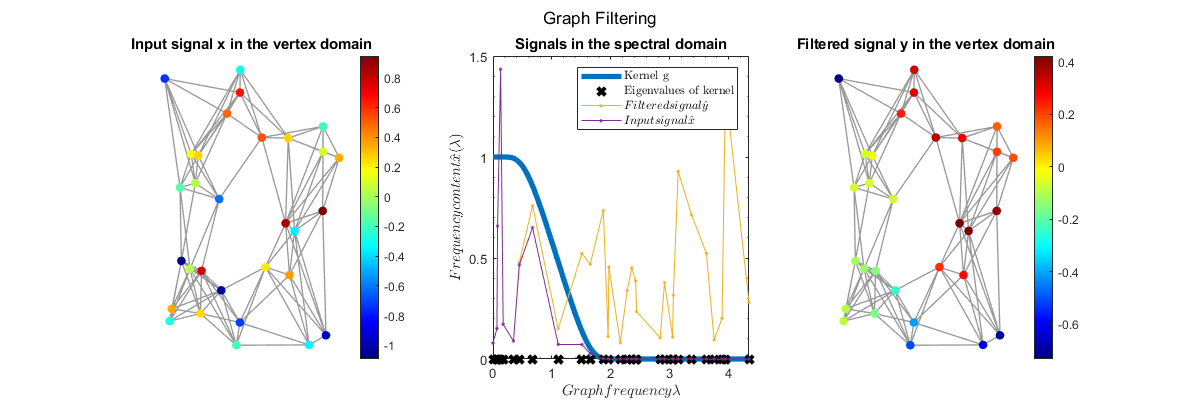} 

    \caption[Caption for LOF]{Effect of a low-pass filter\protect\footnotemark}  

    \label{fig:graph_filter} 

\end{figure} 

\footnotetext{Inspired from \url{https://pygsp.readthedocs.io/en/latest/examples/filtering.html}}

\subsection{application of $S_{DM}$ in graph learning} 

The objective of graph learning is to derive information or relationships about the underlying graph structure given a set of data. A reasonable assumption would be that the data should be smooth so that small changes would be local, based on some understanding of distance, while significant changes would indicate a more considerable distance. Classically, the total variation of a signal is defined as the sum of squared differences in consecutive signal samples $\Sigma _n = (x_n - x_{n-1})^2$, and a signal is called smooth if the quadratic form takes a small value relative to its norm. This concept can be extended to graphs where the notion of neighborhood replaces that of consecutive nodes to obtain 

\[ 
tr(X^{\top} L X) = TV_G (x) = x\top L x = \sum_{n=1}^N \sum_{M \in \mathcal{N}(n)} w_{mn}(x_n - x_m)^2 .  
\] 

This definition of total variation allows us to interpret the ordering of the Laplacian's eigenvalues in terms of frequencies. The larger eigenvalues correspond to higher frequencies (larger total variation). Since, for any function $f \in \mathbb{R}^N$, $f^\top L f = \frac{1}{2} \sum_{i,j = 1}^N w_{ij} (f_i - f_j)^2$, if we let $f$ equal the eigenvector of the graph Laplacian $u_k$, the smoothness $TV(u_k) = u_k^\top L u_k$ gives us the corresponding eigenvalue $\lambda_k$.  

Although there are various methods for learning a graph, we initially adopt the approach proposed in \cite{egilmezGraphLearningFiltered2018}. This method assumes smooth signals and incorporates a pre-filtering step to enhance convergence. The process involves eigendecomposing the sample covariance matrix and applying a graph filter $h(x)$. The output is used to estimate a $L$ that minimizes the total variance. This procedure is repeated iteratively until the maximum likelihood error falls within a specified tolerance.

As an enhancement to the total variation approach, \cite{heimowitzMarkovVariationApproach2020} introduced a Markov measure of variation defined as

\[  
MV(x) = \|x - D^{-1}Wx\| = \|x-Px\|,  
\] 

where a smooth vector implies $\|x-Px\|_1 < f(N)$, with $f$ denoting a function related to the number of nodes. The $l_1$ norm is given by

\[ 
\|x-Px\|_1 = \sum_{i=1}^N \frac{1}{d(v_i)} \left |\sum_{m=1}^N W_{i,m} (x_i - x_m) \right |. 
\] 

Since $P$ can be decomposed into $V\Lambda V^{-1}$ and $\Psi_t = V\Lambda ^t$, the Markov measure of variation is expressed as $MV(x) = |\sum_{i=1}^N \ x_i - \Psi_1^T(i)\hat{x} | < f(N)$.

It is reasonable to assume that the measured signals include distortion or other noise. If we assume that this noise is white, a low-pass filter would reduce the noise while preserving the signal. After the noise is removed, we can analyze the smoothness of the signal to derive information about the underlying graph itself. The process first applies the graph Fourier transform so that the signal is now in the spectral domain $x_g = U^{\top} x$. Next, apply a convolution to the filter, remaining in the spectral domain $(f * g)(i) = \sum_{l=0}^{N-1} \hat{f}(\lambda_l)\hat{g}(\lambda_l)u_l(i) = u_i h(\lambda_i) \hat{x}$. Last, use an inverse graph Fourier transform and bring the result back in the vertex domain $x = u^{-1} \hat{x}$. For this paper, we choose a filter based on Tikhanov Regularization, $h(x) = \frac{1}{1-\tau x}$, where $\tau \in [0,1]$. Since a diffusion map does not uniquely define a graph, we will seek $L$ in the graph learning process, which will require a different representation of $S_{DM}$ as $L$ by $TV = x^{\top}Lx = x^{\top} D^{-1} (P-I) D^{-1} x \approx x^{\top} D^{-1} (\Psi^l -I)D^{-1}x$.  

\begin{figure}[h] 

    \centering 

    \includegraphics[scale=.5]{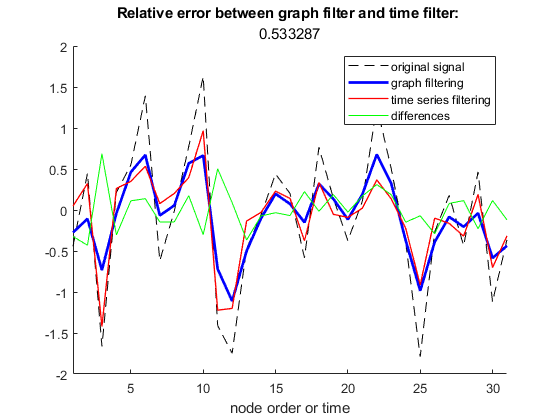} 

    \caption{Comparing graph filters and time filters} 

    \label{fig:relative_errors} 

\end{figure} 

Since the choice of the GSO leads to a specific signal model, the filters and transforms are all based on $S$. Using this framework, we can compare the results of different GSOs. Now, since the filter $H(x)$ smooths the signals and since $\Psi_t(i)$ separates the signal information from noise, using $S_{DM}$ should minimize the oscillation, or total variation, of the signal with respect to the learned graph. Therefore, we have the following:  

\begin{theorem}
    For smooth signals on a connected graph satisfying the following assumptions, the Markov Variation (MV) optimization process converges to a signal with minimal variation and a distribution that approaches the Markov stable distribution of the signals. This convergence is significantly faster than the convergence of Total Variation (TV).
\end{theorem}
Assumptions:

\begin{enumerate}
    \item Connected Graph: The underlying graph $\mathcal{G}$ is connected. This ensures that information can flow freely throughout the graph, which is crucial for accurate diffusion map construction and convergence.
    \item Irreducible Markov Chain: The Markov chain associated with the diffusion map process is irreducible. This means that any state can be reached from any other state with a non-zero probability in a finite number of steps. This property guarantees that the diffusion process can explore the entire graph and capture the global structure of the data.
    \item Finite State Space: The number of states in the Markov chain is finite, corresponding to the number of nodes in the graph. This allows for efficient computation of the diffusion map and ensures that the process reaches a stable state.
    \item Bounded Transition Probabilities: The transition probabilities in the transition matrix $P$ are bounded between 0 and 1. This ensures that the diffusion process reaches a stationary distribution and prevents the signal from exploding or oscillating indefinitely.
    \item Signal Smoothness: The graph signal $x$ is bounded and sufficiently smooth, meaning its second derivative is bounded by a constant $C$. This assumption ensures that the signal variations are gradual and allows for the application of the Rayleigh-Ritz theorem for bounding the difference between TV and MV.
    \item Hermitian Graph Laplacian: The graph Laplacian $L$ is Hermitian, meaning $L=L^\top$. This property guarantees the symmetry of the diffusion process and ensures the validity of certain analytical tools used in the proof.
\end{enumerate}

\begin{proof}

Spectral Properties:
Let $\lambda_1$ and $\lambda_2$ be the first and second smallest eigenvalues of $P$, respectively. Due to connectedness and aperiodicity, $\lambda_1 > 0$ and there exists a positive spectral gap $\lambda_1 - \lambda_2 > 0$. This positive spectral gap is crucial, as it ensures the ergodicity of the diffusion map process and guarantees its convergence to the unique and stable Markov distribution.

Bound on the Difference between TV and MV:
The Rayleigh-Ritz theorem states that for a symmetric matrix $A$ and a real vector $x$, the following inequality holds:
\[
\lambda_{min} \cdot \|x\|^2 \leq x^\top A x \leq \lambda_{max} \cdot \|x\|^2
\]
where $\lambda_{min}$ and $\lambda_{max}$ are the smallest and largest eigenvalues of $A$, respectively.
In the context of this proof, the matrix $A$ is the graph Laplacian $L$, and the vector $x$ represents the graph signal. The leading eigenvector, $\psi_1$, corresponding to the smallest eigenvalue $\lambda_1$, captures the most significant information about the graph structure and encodes the diffusion process. We project the graph signal $x$ onto the leading eigenvector $\psi_1$, obtaining the projection $\hat{x} = \psi_1^\top x$. The smoothness assumption of the signal, which limits the magnitude of its second derivative, allows us to relate the difference between the original signal and its projection to the distance from the leading eigenvector. The spectral gap plays a crucial role in ensuring the convergence and precision of the diffusion map\cite{vonluxburgTutorialSpectralClustering2007}. Using this gap and the smoothness bound, we obtain:
\[
|TV(x) - MV(x)| \leq \sqrt{C} \cdot \sqrt{\lambda_2} \cdot \|x - \Psi(x) \|
\]
where $C$ is the smoothness of the signal and $\Psi=\lambda \psi$ is the diffusion map generated from the signals $x$.

Convergence Rates:
Both TV and MV aim to minimize $\| x - \Psi(x) \|$.
\begin{itemize}
    \item TV: Uses gradient descent, which has an average convergence rate of $O(1/k^2)$ for smooth signals, although the convergence can be slow in the initial stages.
    \item MV: Directly minimizes the distance to the diffusion map representation through projection, resulting in a significantly faster $O(1/k)$ convergence rate due to the spectral gap. This faster rate is achieved because the projection directly aligns the signal with the leading eigenvector, which captures the most significant information in the diffusion process.
\end{itemize}
In general, $O(1/k^2)$ is faster than $O(1/k)$. However, in the context of the proof, the statement is comparing the average convergence rates of two different algorithms. Additionally, the dependence of the bound on the spectral gap explains why MV converges faster than TV. The spectral gap indicates the rate at which irrelevant information decays in the diffusion process, and a larger gap implies faster convergence of MV towards the signal with minimal variation.

Ergodicity and Stable Distribution:
The positive spectral gap guarantees the ergodicity of the diffusion map process. Ergodicity implies that the distribution of the transformed signal converges to a unique and stable Markov distribution, regardless of the initial state of the process. This stable distribution captures the long-term behavior of the diffusion process and reflects the underlying structure of the graph.

In conclusion, MV minimizes the distance to the diffusion map representation, which aligns the signal with the underlying diffusion process. This process inherently minimizes information loss and ensures that signal variations are smooth and consistent with the global structure of the graph. Therefore, the signal obtained through MV optimization can be considered to have minimal variation within the context of the graph and its diffusion process and has a distribution that approaches the Markov stable distribution of the signals. This convergence is significantly faster than the convergence of Total Variation (TV).

\end{proof} 

By minimizing the distance to the diffusion map representation, MV not only reduces the overall variation in the signal, but also promotes its similarity to a signal generated by the Markov process associated with the graph. This means that the signal obtained through MV optimization effectively captures the essential features and dynamics of the diffusion process, making it particularly useful for tasks such as anomaly detection, community detection, and graph signal denoising. We can now incorporate this approach into the graph learning algorithm and measure the accuracy against a known graph and actual data.

%% file: sections/Examples-SDE-DiffusionMap.tex
The general approach for these examples is to first create a graph based on the distance between the nodes and then determine the covariance of the sampled signals. This covariance is then filtered using a low-pass filter based on Tikhonov regularization. The different GSOs are used to construct the filter, and Fourier transforms. Then, the numerical results are compared when the filter parameter $\tau$ is varied and again when the filter order, $t$, is modified. The analysis was done in Matlab, and the graph analysis heavily relied on \cite{GSPBOXmatlab}.

\subsection{Random Sensor Graph} 

\begin{figure}[h] 
    \centering 
    \includegraphics[width=0.75\linewidth]{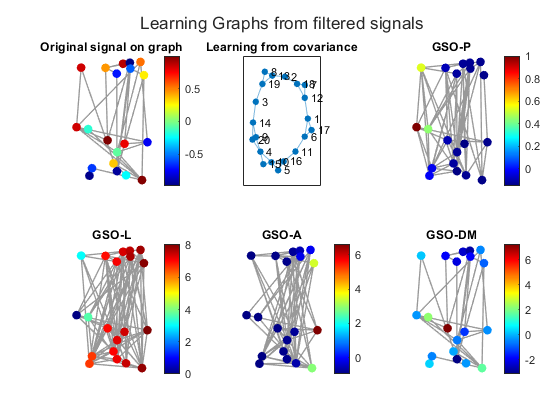} 
    \caption{Learned random sensor graphs} 
    \label{fig:learnedsin}
\end{figure}

\begin{figure}[h] 
    \begin{minipage}{.45\linewidth} 
    \centering 
    \resizebox{\linewidth}{!}{ 
    \includegraphics[scale=.5]{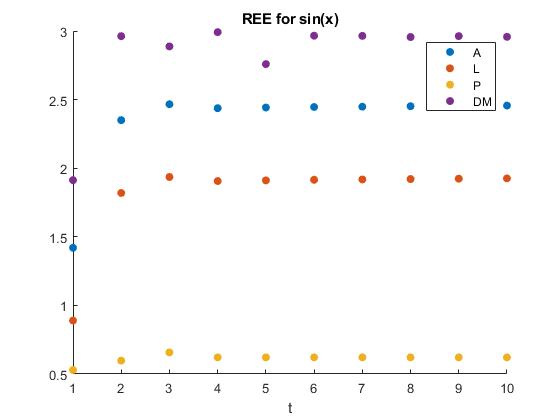} 
    } 
    \caption{Relative Eigenvalue Error - random sensor graph} 
    \label{fig:REEsint} 
    \end{minipage} 
\hspace{0.05\linewidth} % Adjust the space between the tables 
    \begin{minipage}{.45\linewidth} 
    \centering 
    \resizebox{\linewidth}{!}{ 
    \includegraphics[scale=.5]{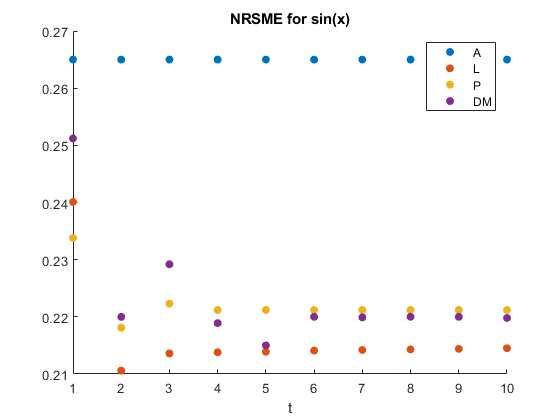} 
    } 
    \caption{NRMSE - random sensor graph varying t} 
    \label{fig:NRMSEsint} 
    \end{minipage} 
\end{figure}

We assess the performance of our proposed graph learning approach by first applying a filter to the sampled covariance, then estimating $L$ by comparing our $S_{DM}$ filter with filters constructed from $A$ and $L$. To do this, we create a sensor graph with $N=100$ nodes and random edges and generate a signal using $sin(x_i)$ that overlays a smooth function from each node. We also maintain a ground truth graph for comparison. We use Tikhonov regularization for all graph filters $h(x) = \frac{1}{1-\tau x}$. We then apply the filter as a convolution and move the signal back to the vertex domain using the inverse GFT. After filtering, the resulting sample covariance matrix is used in the \verb|min_markov_var| function \ref{Appendix}, and the resulting graph is compared with the ground truth graph\ref{fig:learnedsin}. To measure the filtering ability of each GSO, we use the Normalized Root Mean Square Error (NRMSE) $\frac{\sqrt{\frac{1}{n} \sum_{i=1}^{n} (y_i - \hat{y}_i)^2}}{\max(y) - \min(y)}$ and the relative eigenvalue error (REE) given as $\frac{1}{k} \sum_{i=2}^k \frac{ \hat{\lambda_i}-\lambda_i } {\lambda_i}$ between the weighted adjacency matrices of the ground truth graph and the estimated graph. Data with varying $\tau$ are presented in tables \ref{tab:NRSMETableGen} and \ref{tab:REETableGen}, and different $t$ are shown in graphs \ref{fig:REEsint}, \ref{fig:NRMSEsint}, and tables \ref{tab:NRMSEGen-t}, \ref{tab:REEGen-t}. The results show that $S_{DM}$ has a larger REE than the other operators; however, the NRSME levels are similar between the different operators. 

\subsection{Molene Temperature Dataset} 

\begin{figure}[h] 
    \begin{minipage}{.45\linewidth} 
    \centering 
    \resizebox{\linewidth}{!}{ 
    \includegraphics[scale=.5]{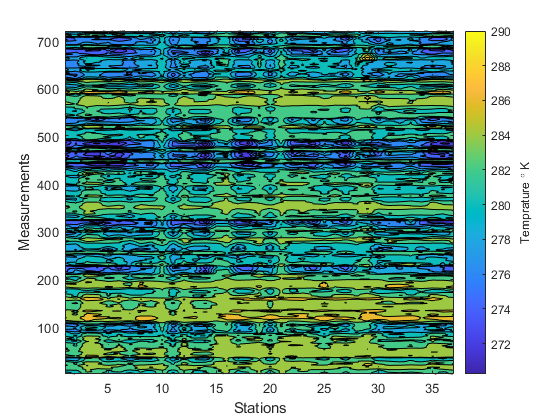} 
    } 
    \caption{Temperature measurements by sensor} 
    \label{fig:temp-surf} 
    \end{minipage} 
\hspace{0.05\linewidth} % Adjust the space between the tables 
    \begin{minipage}{.45\linewidth} 
    \centering 
    \resizebox{\linewidth}{!}{ 
    \includegraphics[scale=.5]{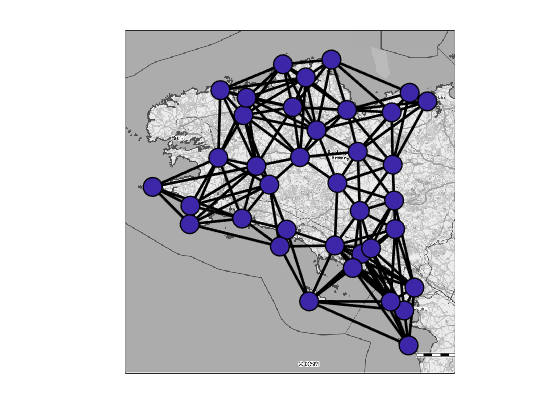} 
    } 
    \caption{Weather Stations, connected by distance} 
    \label{fig:molenedistances} 
    \end{minipage} 
\end{figure}

\begin{figure}[h] 
    \begin{minipage}{.45\linewidth} 
    \centering 
    \resizebox{\linewidth}{!}{ 
    \includegraphics[scale=.5]{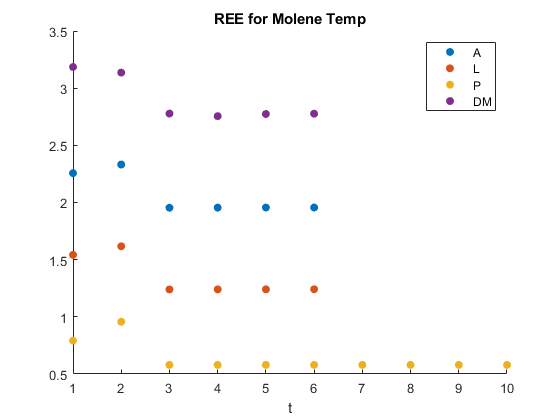} 
    } 
    \caption{Relative Eigenvalue Error - Molene Temperature varying t} 
    \label{fig:REE-Molene} 
    \end{minipage} 
\hspace{0.05\linewidth} % Adjust the space between the tables 
    \begin{minipage}{.45\linewidth} 
    \centering   
    \resizebox{\linewidth}{!}{ 
    \includegraphics[scale=.5]{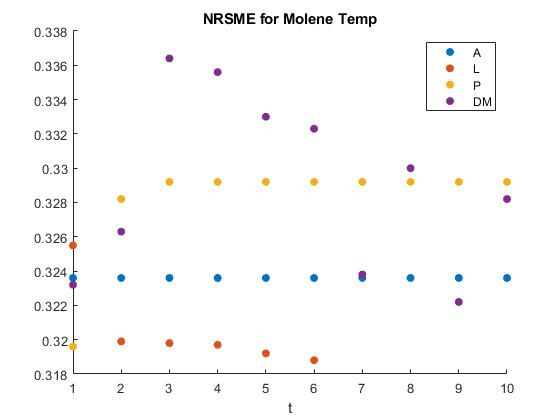} 
    } 
    \caption{NRMSE - Molene Temperature varying t} 
    \label{fig:NRMSE-molene} 
    \end{minipage} 
\end{figure}

\begin{figure}[h] 
    \centering 
    \includegraphics[width=0.75\linewidth]{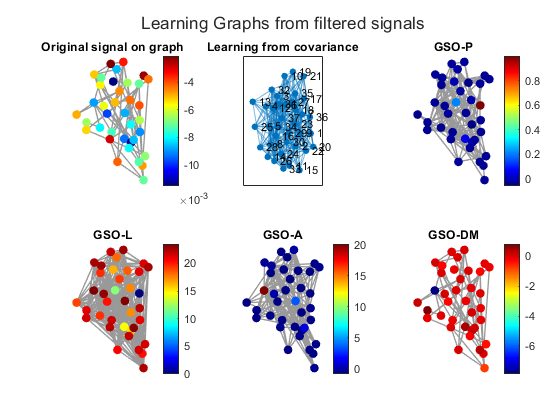} 
    \caption{Learned random sensor graphs} 
    \label{fig:learnedmolene}
\end{figure}

We now analyze a real dataset of hourly temperature measurements taken from N = 37 stations in Brittany, France, in January 2014 \cite{molenedata}. The data includes the sensor locations and a temperature table. A ground-truth graph is created by connecting each station with a radius of 50 km or less \ref{fig:molenedistances}. Signals are the average hourly temperature changes measured for each location. The surface plot \ref{fig:temp-surf} illustrates how the temperature changes are consistent across all stations. We will use different methods to generate the sensor graph and show how temperature measurements can be used to estimate relative sensor locations, even though there is no physical connection. The same filter and graph shift operators are compared, using the exact relative error measurements between the estimated graph and the ground truth graph \ref{fig:learnedmolene}. Data with varying $\tau$ are presented in \ref{tab:NRSMETableMolene} and \ref{tab:REETableMolene} and the effect of changing $t$ in graphs \ref{fig:NRMSE-molene}, \ref{fig:REE-Molene}, and tables \ref{tab:NRSMETableMolene-t}, \ref{tab:REETableMolene-t}. The magnitude of the errors has changed, but the relative differences between $S_{DM}$ and the others remain the same.

\subsection{Comments} 

In both the fabricated and real-world examples, the same errors were observed in the learned graphs. The two error measurements examined the differences between the graphs from different perspectives. The eigenvalue error measured the dissimilarity in eigenvalues between the original and filtered signals, indicating changes in the signal's frequency content. On the other hand, the NRMSE assessed the overall amplitude difference, providing a global measure of the accuracy of the reconstruction. If the filter substantially modifies the spectral content, the eigenvalues of the filtered signal may significantly differ from those of the original signal, resulting in a high relative eigenvalue error. Despite spectral changes, the NRMSE may be lower when the amplitude differences between the original and filtered signals are minor. The filter may maintain the overall shape or trend of the signal while adjusting its frequency components. A diffusion map as a filter utilizes the diffusion process to emphasize certain structural aspects of the data, introducing spatial changes in the graph signal. This can lead to a situation with a high relative eigenvalue error but a lower NRMSE than the original signal, as the diffusion map filter preserves the intrinsic geometry of high-dimensional data, thus resulting in a lower NRMSE despite considerable changes in spectral content. Diffusion maps capture low-frequency components of the graph signal through an iterative diffusion process that smoothes the signal over the graph structure. The number of diffusion steps controls the level of smoothing and the captured frequency range. The diffusion process also smoothes the signal, suppressing high-frequency components and reducing the magnitude of higher eigenvalues of the graph Laplacian matrix. This results in a high relative error in the eigenvalues, particularly for higher ones, even if the overall signal changes are moderate. In addition, diffusion map filtering preserves the low-frequency components that contain most signal energy. The smoothing process reduces the magnitude of local variations, contributing to a lower NRMSE. As in the above examples, diffusion map filtering smoothes the temperature variations, reducing peak temperatures and filling in valleys. This results in a high relative error in higher eigenvalues but a lower NRMSE, as the average temperature remains relatively stable. 
%Lloyd is cool, dad is too.

%% file: sections/Conclusion-SDE-DiffusionMap.tex
\subsection{Conclusion and Contributions}

This paper has delved into the intersection of Diffusion Maps and Graph Signal Processing, focusing on their applications in analyzing and processing complex, non-Euclidean data. Through this exploration, particularly in the context of the Markov Variation problem, our approach has demonstrated significant improvements over other techniques that rely on the Total Variation method. These advances have been effectively applied to a diverse range of data, including both synthetic and real-world temperature sensor datasets. This highlights the versatility of GSP to extend traditional signal processing techniques to irregular graph structures. A key insight of our work is the new perspectives found in preserving the structure of the data while filtering the signals, achieved by extending the signal models using the $S_{DM}$ operator.

A notable limitation of our approach is its reliance on the computation of eigenvectors, which can be inefficient for larger graphs. However, these eigenvectors are not employed directly for graph estimation. Instead, their importance is underscored in the prefiltering phase, which leads to the graph estimation from the covariance of prefiltered signals, done through a regularized maximum likelihood criterion.

\subsection{Open Topics and Future Directions}

Despite the progress made in recent years, several open problems in GSP continue to present opportunities for future research. The impact of graph construction on localized and multiscale transforms is one such area that remains under explored. A deeper understanding of how different graph construction methods affect GSP algorithms is essential for selecting the most effective graph representations for specific problems. Additionally, the criteria for choosing between normalized or nonnormalized graph Laplacian eigenvectors or other bases for spectral filtering are not yet fully clear. This is particularly relevant when considering the extension of GSP methods to more complex graph types, such as infinite or random graphs.

Another area ripe for investigation is the realm of transition probabilities beyond traditional random walks. Exploring the effects of biased walks and jump distributions on GSP is crucial to optimize the use of various probability models in different contexts. Similarly, determining the most suitable distance metric for constructing or analyzing transform methods, whether it is geodesic, resistance, diffusion, or algebraic distances, presents a challenge that requires further exploration.

In addition, adapting GSP methods to dynamic and time-varying graphs is critical to effectively handle evolving real-world data structures. This paper has not only highlighted improvements in extending the signal model using diffusion maps but also sets the stage for future work focused on refining graph estimation methods and exploring new domains where a graph signal approach can be beneficial. The open questions and directions outlined here offer exciting opportunities to further enhance our understanding and analysis of complex, non-Euclidean data.

\clearpage
\pagebreak

%% file: sections/Appendix-SDE-DiffusionMap.tex
\subsection{Bandwidth tuning}

\subsubsection{BGH Epsilon}

This is my Matlab implementation of the BGH estimation algorithm for $\epsilon$.\cite{berryVariableBandwidthDiffusion2016}

\begin{lstlisting}[style=Matlab-editor]
    epsilons = 2.^(-40:40);
    epsilons = sort(epsilons);
    log_T = zeros(size(epsilons));
    for i = 1:length(epsilons)
        eps = epsilons(i);
        log_T(i) =  log(sum(exp((data/4*eps) - max(data/(4*eps), [], "all")), "all")); 
    end
    log_eps = log(epsilons);
    log_deriv = diff(log_T) ./ diff(log_eps);
    [~, max_loc] = max(log_deriv);
    epsilon = exp(log_eps(max_loc))
    d = round(2 * log_deriv(max_loc))
\end{lstlisting}
I didn't find that the method would converge with my data.

\subsubsection{Simple Method for Epsilon}
Instead, I used this simple way to estimate the bandwidth parameter ($\epsilon$) using the median of pairwise distances between data points.

\begin{lstlisting}[style=Matlab-editor]
% Assuming X is the original data matrix
pairwise_distances = pdist(X);
median_distance = median(pairwise_distances);
scale_factor = 0.5;
epsilon = scale_factor * median_distance;
\end{lstlisting}

\subsection{Markov Variation}
Here's my code to minimize the Markov variation. This version is optimized for readability, not speed, and it is quite slow.
\begin{lstlisting}[style=Matlab-editor]
function [est_matrix] = min_markov_var(cov_matrix, n)
    % this function computes an optimal weight matrix, given a sample
    % signal covariance matrix, that minimizes the markov variation |x-Px|

    objective_function = @(W) norm(cov_matrix - pinv(diag(sum(W, 2))) * W * cov_matrix * W', 'fro')^2;
    initial_guess = 0.5 * ones(n); % Initial guess for W

    % Linear inequality constraints (W >= 0)
    A = -eye(n^2);
    b = zeros(n^2, 1);

    % Linear equality constraints (P = D^{-1}W)
    Aeq = [kron(eye(n), ones(1, n)); kron(ones(1, n), eye(n))];
    beq = [ones(n, 1); ones(n, 1)]; % Adjusted the constraints

    % Solve the optimization problem
    options = optimset('Display', 'iter', 'Algorithm', 'interior-point', 'MaxIter', 2000, 'MaxFunEvals', 4000000);
    W_optimal = fmincon(objective_function, initial_guess, A, b, Aeq, beq, [], [], [], options);

    % Make W_optimal symmetric
    W_optimal = 0.5 * (W_optimal + W_optimal.');
    
    est_matrix = W_optimal;
end
\end{lstlisting}
\begin{figure}[h]
    \begin{minipage}{.45\linewidth}
    \centering
    \resizebox{\linewidth}{!}{
    \includegraphics[scale=.5]{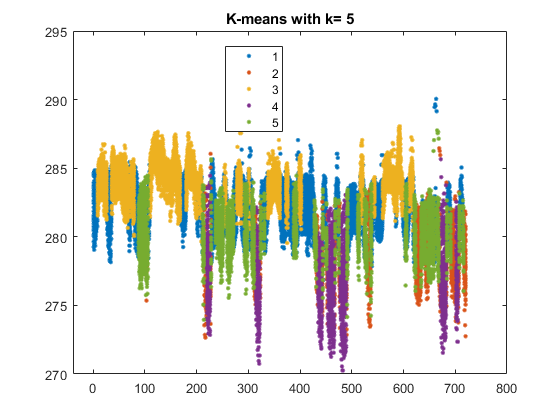}
    }
    \caption{K-means clustering of temp measurements}
    \label{fig:K-meansTemp}
    \end{minipage}
  \hspace{0.05\linewidth} % Adjust the space between the tables
    \begin{minipage}{.45\linewidth}
    \resizebox{\linewidth}{!}{
    \centering
    \includegraphics[scale=.5]{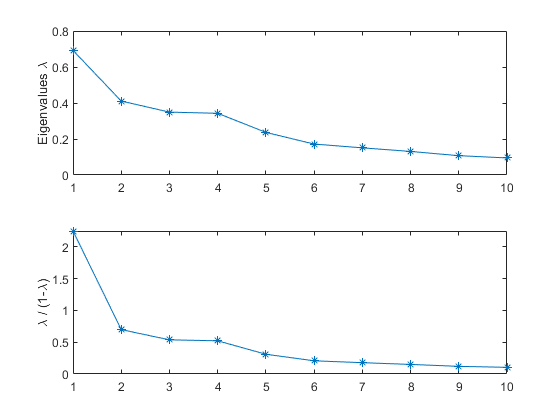}
    }
    \caption{Eigenvalue dropoff for Molene Temp}
    \label{fig:eigendropoff}
    \end{minipage}
\end{figure}

\begin{figure}[h]
    \begin{minipage}{.45\linewidth}
    \centering
    \resizebox{\linewidth}{!}{
    \includegraphics[scale=.5]{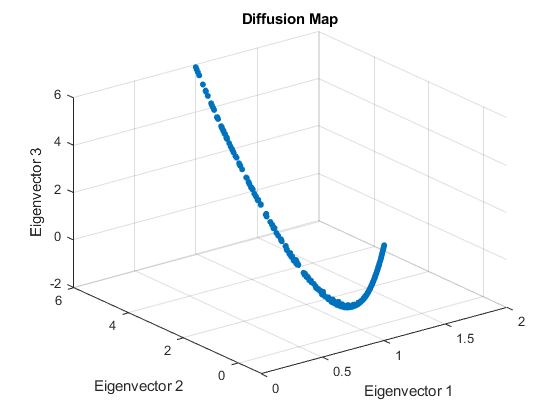}
    }
    \caption{First three Eigenvectors Molene Temp}
    \label{fig:EigenVectorPlot}
    \end{minipage}
\hspace{0.05\linewidth} % Adjust the space between the tables
    \begin{minipage}{.45\linewidth}
    \centering
    \resizebox{\linewidth}{!}{
    \includegraphics[scale=.5]{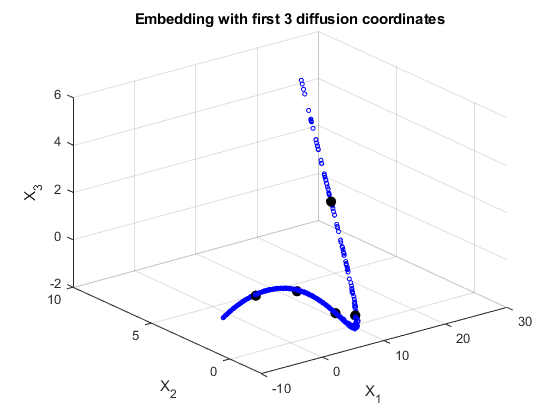}
    }
    \caption{Embedding Molene Temp in lower dimensions}
    \label{fig:embedding}
    \end{minipage}
\end{figure}

\clearpage
\pagebreak

\subsection{Numeric Results from Random Sensor graph}

\begin{table}[h!]
  \centering
  \begin{minipage}{.45\linewidth}
    \centering
    \resizebox{\linewidth}{!}{
    \begin{tabular}{|l||*{5}{c|}}\hline
      \backslashbox{$\tau$}{GSO}
      &\makebox[3em]{A}&\makebox[3em]{L}&\makebox[3em]{P}
      &\makebox[3em]{DM}\\\hline\hline
      0.1 &1.8171&1.2546&0.7395&3.5505\\\hline
      0.2 &1.6086&1.2497&0.7435&2.3608\\\hline
      0.3 &1.8860&1.2173&0.7859&2.6765\\\hline
      0.4 &1.9219&1.1342&0.7500&2.6608\\\hline
      0.5 &1.4519&0.9101&0.5525&1.9507\\\hline
      1.0 &1.4869&1.0597&0.7336&2.1834\\\hline
      1.5 &3.4626&2.8497&1.6614&5.3849\\\hline
      2.0 &2.6277&2.1317&0.8046&3.1984\\\hline
      2.5 &2.1209&1.8520&0.5653&2.7216\\\hline
    \end{tabular}
    }
    \caption{REE - random graph varying $\tau$}
    \label{tab:REETableGen}
  \end{minipage}%
  \hspace{0.05\linewidth} % Adjust the space between the tables
  \begin{minipage}{.45\linewidth}
    \centering
    \resizebox{\linewidth}{!}{
    \begin{tabular}{|l||*{5}{c|}}\hline
      \backslashbox{$\tau$}{GSO}
      &\makebox[3em]{A}&\makebox[3em]{L}&\makebox[3em]{P}
      &\makebox[3em]{DM}\\\hline\hline
      0.1 &0.2594&0.2743&0.2389&0.2798\\\hline
      0.2 &0.2560&0.2637&0.2429&0.2887\\\hline
      0.3 &0.2612&0.2571&0.2457&0.2959\\\hline
      0.4 &0.2402&0.2554&0.2443&0.2898\\\hline
      0.5 &0.2911&0.2570&0.2446&0.2853\\\hline
      1.0 &0.2588&0.2640&0.2810&0.2730\\\hline
      1.5 &0.2714&0.2652&0.2771&0.2705\\\hline
      2.0 &0.2660&0.2654&0.2745&0.2736\\\hline
      2.5 &0.2663&0.2666&0.2576&0.2865\\\hline
    \end{tabular}
    }
    \caption{NRMSE - random graph varying $\tau$}
    \label{tab:NRSMETableGen}
  \end{minipage}

\vspace{1cm} % Adjust the vertical space between the rows of tables

\centering
\begin{minipage}{.45\linewidth}
  \centering
  \resizebox{\linewidth}{!}{
  \begin{tabular}{|l||*{5}{c|}}\hline
    \backslashbox{$t$}{GSO}
    &\makebox[3em]{A}&\makebox[3em]{L}&\makebox[3em]{P}
    &\makebox[3em]{DM}\\\hline\hline
    1 & 1.42 & 0.88937 & 0.52893 & 1.9129 \\\hline
    2 & 2.3502 & 1.8196 & 0.59716 & 2.9625 \\\hline
    3 & 2.4664 & 1.9358 & 0.65721 & 2.8875 \\\hline
    4 & 2.4372 & 1.9066 & 0.62134 & 2.9914 \\\hline
    5 & 2.442 & 1.9114 & 0.62134 & 2.7588 \\\hline
    10 & 2.4566 & 1.926 & 0.62134 & 2.9576 \\\hline
  \end{tabular}
  }
  \caption{REE random graph varying t}
  \label{tab:REEGen-t}
\end{minipage}
\hspace{0.05\linewidth} % Adjust the space between the tables
\begin{minipage}{.45\linewidth}
  \centering
  \resizebox{\linewidth}{!}{
  \begin{tabular}{|l||*{5}{c|}}\hline
    \backslashbox{$t$}{GSO}
    &\makebox[3em]{A}&\makebox[3em]{L}&\makebox[3em]{P}
    &\makebox[3em]{DM}\\\hline\hline
    1 & 0.265 & 0.2401 & 0.2338 & 0.2512 \\\hline
    2 & 0.265 & 0.2106 & 0.2181 & 0.22 \\\hline
    3 & 0.265 & 0.2136 & 0.2223 & 0.2292 \\\hline
    4 & 0.265 & 0.2138 & 0.2212 & 0.2189 \\\hline
    5 & 0.265 & 0.2139 & 0.2212 & 0.215 \\\hline
    10 & 0.265 & 0.2145 & 0.2212 & 0.2198 \\\hline
  \end{tabular}
  }
  \caption{NRMSE - random graph varying t}
  \label{tab:NRMSEGen-t}
\end{minipage}
\end{table}

\subsection{Numeric Results from Molene Temperature Data}

\begin{table}[h!]
  \centering
  \begin{minipage}{.45\linewidth}
    \centering
    \resizebox{\linewidth}{!}{
    \begin{tabular}{|l||*{5}{c|}}\hline
      \backslashbox{$\tau$}{GSO}
      &\makebox[3em]{A}&\makebox[3em]{L}&\makebox[3em]{P}
      &\makebox[3em]{DM}\\\hline\hline
      0.1&2.4664&1.7204&0.9360&3.2280\\\hline
      0.2&2.3366&1.6578&0.9086&3.2364\\\hline
      0.3&2.4262&1.6746&0.9261&3.1886\\\hline
      0.4&2.2936&1.6242&0.8984&3.1021\\\hline
      0.5&2.2569&1.5420&0.7922&3.1850\\\hline
      0.6&2.3027&1.6378&0.8782&3.1811\\\hline
      0.7&2.4425&1.6226&0.8631&3.0221\\\hline
      0.8&2.3752&1.6078&0.8468&2.9512\\\hline
      0.9&2.3079&1.5293&0.7696&3.0209\\\hline
      1.0&2.4661&1.6386&0.8880&3.3743\\\hline
    \end{tabular}
    }
    \caption{REE - Molene Temp varying $\tau$}
    \label{tab:REETableMolene}
  \end{minipage}%
  \hspace{0.05\linewidth} % Adjust the space between the tables
  \begin{minipage}{.45\linewidth}
    \centering
    \resizebox{\linewidth}{!}{
    \begin{tabular}{|l||*{5}{c|}}\hline
      \backslashbox{$\tau$}{GSO}
      &\makebox[3em]{A}&\makebox[3em]{L}&\makebox[3em]{P}
      &\makebox[3em]{DM}\\\hline\hline
      0.1&0.3176&0.3249&0.3235&0.3300\\\hline
      0.2&0.3182&0.3255&0.3188&0.3228\\\hline
      0.3&0.3250&0.3255&0.3293&0.3258\\\hline
      0.4&0.3220&0.3255&0.3195&0.3309\\\hline
      0.5&0.3236&0.3255&0.3196&0.3232\\\hline
      0.6&0.3227&0.3255&0.3241&0.3196\\\hline
      0.7&0.3268&0.3256&0.3167&0.3230\\\hline
      0.8&0.3249&0.3256&0.3148&0.3282\\\hline
      0.9&0.3245&0.3255&0.3241&0.3284\\\hline
      1.0&0.3248&0.3252&0.3155&0.3282\\\hline
    \end{tabular}
    }
    \caption{NRMSE - Molene Temp varying $\tau$}
    \label{tab:NRSMETableMolene}
  \end{minipage}
  
\vspace{1cm} % Adjust the vertical space between the rows of tables

  \centering
  \begin{minipage}{.45\linewidth}
    \centering
    \resizebox{\linewidth}{!}{
    \begin{tabular}{|l||*{5}{c|}}\hline
      \backslashbox{$t$}{GSO}
      &\makebox[3em]{A}&\makebox[3em]{L}&\makebox[3em]{P}
      &\makebox[3em]{DM}\\\hline\hline
      1 & 2.2569 & 1.542 & 0.7922 & 3.185 \\\hline
      2 & 2.3322 & 1.6172 & 0.9568 & 3.136 \\\hline
      3 & 1.9544 & 1.2395 & 0.5796 & 2.7777 \\\hline
      4 & 1.9552 & 1.2402 & 0.5796 & 2.7549 \\\hline
      5 & 1.9558 & 1.2409 & 0.5796 & 2.774 \\\hline
      10 & NaN & NaN & 0.5796 & NaN \\\hline
    \end{tabular}
    }
    \caption{REE - Molene Temp varying $t$}
    \label{tab:REETableMolene-t}
  \end{minipage}%
  \hspace{0.05\linewidth} % Adjust the space between the tables
  \begin{minipage}{.45\linewidth}
    \centering
    \resizebox{\linewidth}{!}{
    \begin{tabular}{|l||*{5}{c|}}\hline
      \backslashbox{$t$}{GSO}
      &\makebox[3em]{A}&\makebox[3em]{L}&\makebox[3em]{P}
      &\makebox[3em]{DM}\\\hline\hline
      1 & 0.3236 & 0.3255 & 0.3196 & 0.3232 \\\hline
      2 & 0.3236 & 0.3199 & 0.3282 & 0.3263 \\\hline
      3 & 0.3236 & 0.3198 & 0.3292 & 0.3364 \\\hline
      4 & 0.3236 & 0.3197 & 0.3292 & 0.3356 \\\hline
      5 & 0.3236 & 0.3192 & 0.3292 & 0.3330 \\\hline
      10 & 0.3236 & NaN & 0.3292 & 0.3282 \\\hline
    \end{tabular}
    }
    \caption{NRMSE - Molene Temp varying $t$}
    \label{tab:NRSMETableMolene-t}
  \end{minipage}
\end{table}